\documentclass{article}

\usepackage{arxiv}

\usepackage[utf8]{inputenc} 
\usepackage[T1]{fontenc}    
\usepackage{hyperref}       
\usepackage{url}            
\usepackage{booktabs}       
\usepackage{amsfonts}       
\usepackage{nicefrac}       
\usepackage{microtype}      
\usepackage{lipsum}		
\usepackage{graphicx}
\usepackage{natbib}
\usepackage{doi}
\usepackage{tikz}
\usepackage{booktabs}  
\usepackage{array}     
\usepackage{threeparttable}  
\usepackage{amsmath}      
\usepackage{amssymb}      
\usepackage{tikz}
\usepackage{pgfplots}
\pgfplotsset{compat=1.18}
\usetikzlibrary{trees}
\usepackage{tikz}
\usetikzlibrary{arrows.meta, positioning, shapes.multipart, fit, calc}
\usepackage{tikz}
\usetikzlibrary{shapes.geometric, arrows.meta, positioning, fit, backgrounds, calc, shapes.symbols}
\usepackage[ruled,vlined]{algorithm2e}
\usepackage{amsmath}
\usepackage{float}

\title{A centroid based framework for text classification in itsm environments}

\author{%
  \href{https://www.linkedin.com/in/hosein-mohanna/}{\includegraphics[scale=0.06]{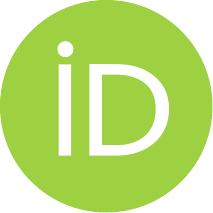}\hspace{1mm}Hossein MOHANNA}\thanks{Further information about author (webpage, alternative address)} \\
  Global AI Lab\\
  EasyVista\\
  \texttt{hmohanna@easyvista.com}
  \And
  \href{https://orcid.org/0009-0004-2715-7363}
  {\includegraphics[scale=0.06]{orcid.pdf}\hspace{1mm}Ali AIT BACHIR}\thanks{Further information about author (webpage, alternative address)} \\
  Global AI Lab\\
  EasyVista\\
  \texttt{aaitbachir@easyvista.com}
}

\date{October 2025}

\renewcommand{\headeright}{}
\renewcommand{\undertitle}{}
\renewcommand{\shorttitle}{A centroid based framework for text classification in itsm environments}

\hypersetup{
pdftitle={A centroid based framework for text classification in itsm environments},
pdfsubject={AIOps, embedding, centroid, transformer, categorization},
pdfauthor={Hossein MOHANNA},
pdfkeywords={AIOps, embedding, centroid, transformer, categorization},
}

\begin{document}
\maketitle

\begin{abstract}
Text classification with hierarchical taxonomies is a fundamental requirement in IT Service Management (ITSM) systems, where support tickets must be categorized into tree-structured taxonomies. We present a dual-embedding centroid-based classification framework that maintains separate semantic and lexical centroid representations per category, combining them through reciprocal rank fusion at inference time. The framework achieves performance competitive with Support Vector Machines (hierarchical F1: 0.731 vs 0.727) while providing interpretability through centroid representations. Evaluated on 8,968 ITSM tickets across 123 categories, this method achieves 5.9 times faster training and up to 152 times faster incremental updates. With 8.6-8.8 times speedup across batch sizes (100-1000 samples) when excluding embedding computation. These results make the method suitable for production ITSM environments prioritizing interpretability and operational efficiency.
\end{abstract}


\section{Introduction}
\label{sec:introduction}

Automated text classification is a fundamental component of modern information systems. When documents must be categorized into hierarchical taxonomies, such as tree-structured label spaces, classification systems can leverage parent-child relationships between categories to provide better predictions. Hierarchical Text Classification (HTC) addresses this scenario by considering taxonomic relationships during model training and inference ~\cite{Zangari2024b}.

Ticket categorization or classification is a fundamental problem in Information Technology Service Management (ITSM), where systems process thousands of support tickets every day. These tickets must be categorized accurately and promptly in the correct hierarchy to facilitate triage, routing, and resolution. While numerous methods for classification have been developed in the research community, including both traditional machine learning and deep learning approaches, their applicability in real-world enterprise environments remains limited. A significant gap exists between the capabilities of these models under controlled academic conditions and the practical demands of enterprise-grade ITSM systems.

ITSM datasets present further complications. They can have hundreds of categories, and are often imbalanced with some categories having thousands of samples while others contain only a handful. The hierarchy itself may be irregular, with some branches extending several levels deep while others remain shallow. Moreover, enterprise taxonomies are dynamic, frequently evolving as business needs change. Categories may be added, removed, or reorganized over time, and classification systems must be able to adapt accordingly, ideally without requiring full retraining. Inference systems must also be always-ready, avoiding costly delays associated with model loading.

Beyond these challenges, ITSM data exhibits structural properties that diverge significantly from those assumed in academic benchmarks. While most HTC methods assume that classification targets are located at the leaf nodes of the hierarchy, enterprise tickets often have valid labels at intermediate levels. For instance, an issue labeled as “Infrastructure/Network” may be complete and sufficient for resolution, without the need for deeper subcategory assignments ~\cite{wang2024leveraging, plaud2024revisiting}. Methods that force classification to the deepest node may therefore misrepresent or over-specify the intended label.

Another critical difference lies in the semantic structure of the label space. In many ITSM taxonomies, labels are not meaningful phrases but system-generated identifiers such as “Category\_001” or “INCTYPE\_47.” This makes it infeasible to rely on techniques that extract or infer meaning from label names or use semantic similarity to guide classification. Some current approaches rely on semantic relationships between labels ~\cite{chen2023enhancing}, which are unavailable when labels lack semantic content.

Considering these challenges, we present a dual-embedding centroid-based classification framework that addresses the practical requirements of ITSM deployment. Our approach maintains separate semantic (SBERT) and lexical (TF-IDF) centroid representations per category, combining them via reciprocal rank fusion at inference time. The framework provides configurable scoring strategies that can be selected based on dataset characteristics through empirical validation. Evaluated on real-world ITSM tickets, our method achieves performance competitive with Support Vector Machines while providing interpretability,  faster training, and faster incremental updates which are critical properties for production environments with evolving taxonomies and interpretability requirements.

The remainder of this paper is organized as follows. Section~\ref{sec:related-work} reviews relevant work on HTC, centroid-based methods, and text classification. Section~\ref{sec:contribution} presents our dual-embedding framework architecture, configurable scoring strategies, and algorithmic details. Section~\ref{sec:experimental-setup} describes the dataset, preprocessing, and evaluation methodology. Section~\ref{sec:results} reports experimental results and computational efficiency analysis. Section~\ref{sec:conclusion} concludes with future research directions.

\section{Related work}
\label{sec:related-work}

Density-based clustering with document embeddings has been explored for unsupervised text categorization~\cite{radulescu2020density}, demonstrating feasibility but exposing fundamental challenges in scalability and cluster alignment when applied to high-dimensional data. Consequently, while significant advances have been made in representation learning and clustering, there is a clear need for domain-adapted solutions tailored to the structure and complexity of ITSM data. ~\cite{wahba2020evaluating} compared static word embedding models with TF–IDF representation on 1.6 million support tickets spanning 32 categories. They showed that TF–IDF combined with a linear SVM matched or even outperformed the embedding models while keeping training time and computation cost low.  The authors attribute this result to two characteristics of ITSM text, that many domain-specific terms are out-of-vocabulary for off-the-shelf embeddings, and technical language contains little polysemy, reducing the advantage of distributed representations.  ~\cite{revina2020it} reached a similar conclusion, reporting that an \(n\)-gram TF–IDF model combined with logistic regression performs better on IT-ticket classification tasks, mentioning the “the simpler, the better”. ~\cite{alhawari2021machine} built a help-desk prototype that vectorised ticket text with TF–IDF and routed requests via a linear SVM, demonstrating substantial reductions in manual effort on real enterprise data. Similarly, recent empirical studies comparing Naive Bayes and Support Vector Machine (SVM) methods for helpdesk ticket classification reaffirm SVM’s superior accuracy over Naive Bayes ~\cite{wibowo2023comparison}. The literature review by ~\cite{fuchs2022improving} synthesized findings from primary studies and confirmed that linear SVM and Random Forest remain the most consistently strong performers for support-ticket classification, while forecasting that deep-learning architectures can be superior in the future. While transformer models show promise, recent work demonstrates that combining complementary representations through simple fusion methods can be highly effective. ~\cite{esteva2021covid} showed that fusing SBERT semantic search with traditional BM25 using Reciprocal Rank Fusion significantly outperformed single-view approaches in their search system.


Transformer-based models have revolutionized natural language processing by enabling powerful contextualized representations of text. Since the introduction of the Transformer architecture~\cite{vaswani2017attention}, models such as BERT~\cite{devlin2019bert} and its variants have achieved state-of-the-art performance across a wide range of tasks including text classification. Traditional keyword matching approaches, word embeddings or TF-IDF can fail to capture the true intent behind support requests in ITSM ticket classification. Unlike word embeddings, sentence embeddings using pretrained embedding models capture the broader semantic context of entire phrase making them particularly valuable for understanding complex ticket descriptions in ITSM environments. Multilingual embedding models ensure that similar content in different languages maintains proximity in the vector space ~\cite{reimers2019sentence}~\cite{reimers2020making}.

~\cite{galke2023we} revisits the question “Are we really making much progress in text classification?” by comparing flat single-label, flat multi-label and hierarchical text classification. They show that fine-tuning encoder-only transformers such as BERT especially hierarchy-aware variants achieves strong results, yet simple BoW models remain hard to beat. TF-IDF combined with a Wide-MLP, logistic regression or trigram SVM still provides good baselines. From a performance perspective, even advanced pre-trained models like BERT or XLNet have shown limited effectiveness in IT ticket classification. ~\cite{Zangari2023} reported F1-scores between 38\% and 55\% on real-world datasets. The study attributed this to the domain gap between general-purpose language models and the technical jargon and formatting inconsistencies typical of IT tickets. Hierarchical classification, while promising in structured contexts, was less effective when applied to noisy or inconsistently labeled data. Moreover, a recent benchmark study also reports a negative correlation between model complexity and F1 score for datasets smaller than 10 000 instances. Concluding that expensive transformers can be unnecessary and simple algorithms such as logistic regression can rival deep architectures ~\cite{reusens2024evaluating}. These are empirically validated by recent studies on enterprise ITSM data. ~\cite{wahba2022comparison} found that even fine-tuned Pre-trained Language Models (BERT, RoBERTa, DistilBERT, XLM) failed to significantly outperform linear SVM classifiers on real IT support tickets. Regarding LLMs, ~\cite{cunha2025thorough} provides compelling empirical evidence, showing that while LLMs achieve higher effectiveness, they are 590 times slower than traditional methods and 8.5 times slower than small language models, making them impractical for high-volume enterprise environments. Although recent LLMs and few-shot methods appear promising for HTC, their high cost and limited context make them unattractive for production-level ITSM ticket classification. LLMs with sophisticated prompts can still lag behind well-tuned smaller transformers. However, in our current ITSM setting even these smaller fine-tuned SLMs, and certainly large LLMs, are impractical. As they violate strict on-prem data-privacy requirements when offered only via cloud APIs, and demand GPU resources and memory footprints unavailable in typical enterprise servers. They also introduce unacceptable inference latency under high ticket volumes, require specialised ML expertise for continual re-training as the hierarchy evolves, and provide limited interpretability.


Recent work frames Hierarchical Text Classification (HTC) as a special case of classification in which document labels form a tree or DAG-structured taxonomy; exploiting these dependencies is key to a better performance than flat text classifiers that ignore the hierarchy ~\cite{Zangari2024b}. They categorise existing methods into (i) flat models that discard the hierarchy and treat only the leaves, (ii) local models that train many smaller binary or multiclass classifiers on sub-parts of the hierarchy, and (iii) global models that learn a single model for the entire taxonomy. Local approaches require top-down inference and therefore suffer from error propagation, once a parent is mis-classified, all its descendants become unreachable. The survey also distinguishes non-mandatory leaf (NMLNP) settings common in ITSM ticket routing where stopping at an internal node is acceptable, from mandatory leaf (MLNP) scenarios. They also argue that flat metrics are not suited for HTC especially in NMLNP scenarios because they ignore the severity of mistakes along the hierarchy, recommending hierarchy-aware metrics.


Centroid-based classification represents one of the most intuitive approaches to classification, where each class is characterized by a vector computed as the mean of all training embeddings assigned to that class~\cite{han2000centroid}. Classification proceeds by assigning new sample to the class with the most similar centroid using some form of similarity such as cosine. The Hierarchical Centroid-Based Classifier System (HCCS) is an approach to centroid-based hierarchical classification applicable to text or any other vector space~\cite{ferrandin2015centroid}. This method adapts traditional centroid-based classification techniques to handle hierarchical taxonomies by creating class centroids through simple averaging of TF-IDF vectors. What distinguishes HCCS from flat centroid-based methods is its hierarchical training strategy, where each training instance contributes not only to its directly assigned class centroid but also to all ancestor class centroids in the taxonomy. This propagation mechanism ensures that parent class centroids capture the collective characteristics of their descendant classes. During classification, the system computes cosine similarity between test instances and all class centroids, assigning instances to the class with the highest similarity score. Label assignment in HCCS follows the SPL (Single Path of Labels) paradigm with NMLNP, so each instance is mapped to one root-to-node path and the prediction may legitimately stop at an internal class rather than being forced to reach a leaf. However, HCCS exhibits several limitations such as its reliance on TF-IDF vectors alone which limits semantic relationship capture (especially in multilingual setting). The single centroid per class constraint may inadequately represent classes with diverse subclusters, and the ancestor propagation strategy lacks flexibility. \newline

Centroid-based approaches offer significant computational advantages. Training requires only computing and storing category centroids, while inference involves simple distance calculations between query embeddings and the stored centroids. Unlike complex architectures, centroid-based methods provide clear explanations for their predictions. The similarity scores directly indicate how closely a document resembles each category, and the centroid composition reveals which training documents most influence the category representation. This approach also scales well. Adding new categories merely involves computing their centroids without retraining the entire model, a property that proves particularly valuable in enterprise environments where taxonomies evolve frequently. However, simple centroid-based methods face limitations when dealing with categories that exhibit high internal variance. In such cases, a single centroid may fail to capture the diversity of the documents it represents, resulting in lower classification accuracy. Beyond position-based centroid adjustments, other approaches explore weighting strategies to address class imbalance in centroid-based classification ~\cite{liu2017new}. However, these improvements focus on flat classification and do not address the dual challenges of hierarchical taxonomy navigation and multi-representation fusion. While hierarchical centroid-based methods like HCCS offer strong baselines, their reliance on single embeddings limits their ability to leverage complementary semantic and lexical information. Our work extends these approaches by combining dual embeddings through RRF fusion.

\section{Methodology}
\label{sec:contribution}

\subsection{Problem Formalization}
\label{sec:formalization}

Hierarchical Text Classification (HTC) extends traditional flat classification by considering the label hierarchy. In HTC, documents are classified into categories organized in a tree or graph structure where broader categories contain more specific subcategories. For example, in ITSM systems, a ticket might be classified as "Incident/Hardware/Server/Memory Failure" where each level represents increasing specificity. The fundamental characteristic that distinguishes HTC from flat classification is the hierarchy constraint. If a document belongs to a specific category, it must also belong to all parent categories in the hierarchy~\cite{Zangari2024b}.

HTC encompasses several distinct problem formulations that differ in their prediction requirements. In single-label prediction, each document receives exactly one classification path from the root to a leaf node. While multi-label prediction is where documents can belong to multiple categories simultaneously. However, our approach focuses primarily on top-k prediction where the system returns the k most likely category paths without requiring a single definitive answer. This best serves the practical needs of ITSM environments where support agents benefit from multiple relevant suggestions rather than rigid single-category automatic assignments that may be incorrect.

Traditional centroid-based approaches offer computational efficiency and interpretability but face limitations when applied to hierarchical classification. Flat centroid approaches treat parent and child categories as independent entities, discarding taxonomic relationships. Moreover, their reliance on one embedding type may miss complementary semantic or lexical information. Finally, standard similarity scoring does not account for hierarchical paths or adapt to dataset depth distributions.

\subsection{Framework}
\label{sec:our-method}

We present a dual-embedding centroid framework that addresses these limitations while preserving the computational and interpretability advantages of centroid-based methods. Unlike traditional approaches that rely on a single embedding type or concatenate features during training, our method maintains separate semantic and lexical centroid representations per category and fuses them at inference time through reciprocal rank fusion (RRF). This dual representation addresses the complementary strengths of different embedding approaches, particularly valuable in ITSM environments where both semantic understanding and precise terminology matter.

\begin{figure}[t]
\centering
\begin{tikzpicture}[
    node distance=1cm,
    process/.style={rectangle, rounded corners, draw=black!60, fill=blue!15, 
                    thick, minimum width=4.5cm, minimum height=0.8cm, align=center},
    io/.style={trapezium, trapezium left angle=70, trapezium right angle=110,
               draw=black!60, fill=green!15, thick, minimum width=4cm, 
               minimum height=0.8cm, align=center},
    parallel/.style={rectangle, draw=purple!60, fill=purple!10, 
                     thick, minimum width=3.5cm, minimum height=0.8cm, align=center},
    arrow/.style={-{Stealth[length=2.5mm]}, thin}
]

\node[io] (input) {\textbf{Input:} Query text};

\node[process, below=of input] (step1) {Generate Embeddings};

\node[parallel, below=1.25cm of step1, xshift=-2.5cm] (rank1) {SBERT Path Scoring\\(primary centroids)};
\node[parallel, below=1.25cm of step1, xshift=2.5cm] (rank2) {TF-IDF Path Scoring\\(secondary centroids)};

\node[font=\small\bfseries, below=1.3cm of step1] (parallel_label) {Ranking};

\node[process, below=of rank1] (list1) {Rank list $R_{\text{SBERT}}$};
\node[process, below=of rank2] (list2) {Rank list $R_{\text{TF-IDF}}$};

\node[process, below=1cm of step1, yshift=-4cm] (rrf) {Reciprocal Rank Fusion};

\node[io, below=of rrf] (output) {\textbf{Output:} Top-$k$ category predictions};

\draw[arrow] (input) -- (step1);
\draw[arrow] (step1) -- ++(0,-0.8) -| (rank1);
\draw[arrow] (step1) -- ++(0,-0.8) -| (rank2);
\draw[arrow] (rank1) -- (list1);
\draw[arrow] (rank2) -- (list2);
\draw[arrow] (list1) |- (rrf);
\draw[arrow] (list2) |- (rrf);
\draw[arrow] (rrf) -- (output);

\node[draw=purple!60, dashed, thick, rounded corners, 
      fit=(rank1)(rank2)(list1)(list2), inner sep=0.3cm] {};

\end{tikzpicture}
\caption{Inference phase with parallel dual-ranking and RRF fusion.}
\label{fig:inference_flow}
\end{figure}
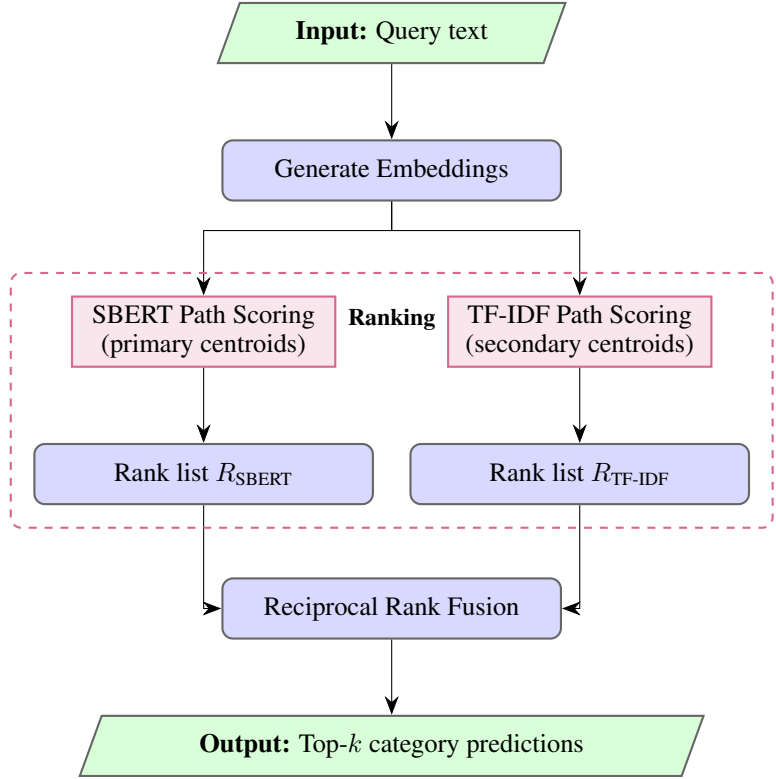

The framework can extend to automatic multi-centroid clustering for categories with high intra-category variance. Categories exceeding a configurable threshold of samples can be partitioned into $k \in [2, \text{max\_clusters}]$ subclusters using methods such as agglomerative clustering, with the optimal $k$ selected to maximize average silhouette score. For our dataset, empirical evaluation showed single centroids achieve competitive performance while maximizing computational efficiency, so multi-centroid clustering was disabled. Centroid computation follows a post-order traversal of the hierarchy tree, ensuring that child node information is available when processing parent nodes. The method supports two strategies for incorporating hierarchical structure. Such as direct instance aggregation where all training documents directly assigned to a category contribute to its centroid computation. And optional hierarchical child sampling where the method can sample a configurable proportion of instances from child categories to ensure parent centroids reflect their subtaxonomy. For depth-homogeneous hierarchies like our ITSM dataset (90.8\% at depth 3), this sampling provides minimal benefit and is disabled. This approach would address the stale node problem where intermediate categories lack direct training data while preserving hierarchical relationships.

\begin{algorithm}[H]
\SetAlgoLined
\caption{Centroid Prediction}
\label{alg:inference_final}

\KwIn{Query text, trained hierarchy tree, scoring strategy}
\KwOut{Top-$k$ predicted categories}

\BlankLine
\tcp{Encode query with both methods}
Encode query using SBERT and TF-IDF\;

\BlankLine
\tcp{Score all category paths}
\ForEach{category path $p$ in tree}{
    \tcp{For each embedding type independently}
    \ForEach{node in path $p$}{
        Compute max cosine similarity with centroids at node\;
    }
    
    Apply scoring strategy to compute path score\;
}

\BlankLine
\tcp{Fuse the two independent rankings}
Rank all paths by SBERT scores\;
Rank all paths by TF-IDF scores\;

\ForEach{path $p$}{
    Get rank positions from both rankings\;
    Compute RRF score from both rank positions\;
}

\BlankLine
\Return{Top-$k$ paths with highest RRF scores}\;

\end{algorithm}

For a query document, we generate two independent rankings from the two embedding types. To combine rankings from multiple embedding types (classification), we employ reciprocal rank fusion (RRF)~\cite{cormack2009reciprocal}, a proven rank aggregation method that effectively merges ranked lists from different retrieval systems ~\cite{rackauckas2024rag}.

For a query document with embedding $q$ and a category path $P = [n_1, n_2, ..., n_{\text{leaf}}]$ 
from root to leaf, we compute a path score by aggregating node-level similarities. Each node $n_i$ 
stores a set of centroids, and we compute the maximum cosine similarity $\cos(q, c)$ between the 
query embedding and any centroid at that node. The path scoring strategy determines how these 
node-level similarities are combined into a final path score. Rather than imposing a fixed scoring strategy, different scoring options can be used which can adapt to the data characteristics. We propose the following three scoring strategies as baselines: 

\textbf{Leaf-Only Scoring} : Uses only the most specific category (leaf node) for classification.
\begin{equation}
\text{PathScore}(q, P) = \max_{c \in \text{centroids}(n_{\text{leaf}})} \cos(q, c)
\end{equation}

\textbf{Simple Average} (equal node weights): Considers all nodes along the hierarchical path with equal weight. This treats all hierarchy levels equally, averaging similarities from root to leaf.
\begin{equation}
\text{PathScore}(q, P) = \frac{1}{|P|} \sum_{n_i \in P} \max_{c \in \text{centroids}(n_i)} \cos(q, c)
\end{equation}

\textbf{Weighted Scoring} Similar to average-based but applies higher weights to deeper nodes, to emphasize deeper and more specific nodes. 

Scoring strategy selection can be automated via validation set performance or manually configured based on hierarchy depth distribution analysis. Our empirical evaluation demonstrates that the optimal strategy depends on dataset characteristics.

\begin{algorithm}[H]
\SetAlgoLined
\caption{Centroid Training}
\label{alg:training_final}

\KwIn{Training samples with hierarchical category labels}
\KwOut{Category tree with dual centroids at each node}

\BlankLine
\tcp{Build category hierarchy}
\ForEach{sample in training data}{
    Insert sample into tree following its category path\;
}

\BlankLine
\tcp{Encode all samples with both methods}
Encode all samples using SBERT and TF-IDF\;

\BlankLine
\tcp{Compute centroids bottom-up}
\ForEach{node $n$ in tree (leaves first, then parents)}{
    \tcp{Collect embeddings for this node}
    \eIf{node has direct training samples}{
        Collect embeddings from node's samples\;
    }{
        \tcp{Stale node: inherit from children}
        Aggregate all embeddings from child nodes\;
    }
    
    \tcp{Create centroid representations}
    \eIf{multi-centroid clustering is enabled}{
        Cluster embeddings into optimal number of centroids\;
    }{
        Compute centroid as mean of embeddings\;
    }
    
    Store centroids for both embedding types at node $n$\;
}

\end{algorithm}

Our dual-embedding centroid framework addresses practical ITSM deployment requirements through four core principles:

\textbf{Dual-Embedding Architecture with RRF Fusion:}  
Unlike traditional centroid methods using single embedding types, our framework maintains separate semantic (SBERT) and lexical (TF-IDF) centroid representations. Independent rankings from each embedding type are fused via reciprocal rank fusion, leveraging complementary strengths.

\textbf{Configurable Scoring Strategies:}  
The framework can integrate different scoring options that can adapt to dataset characteristics. Strategy selection can be automated via validation set evaluation or manually configured based on hierarchy depth distribution, demonstrating that optimal architecture depends on empirical data structure rather than prescriptive assumptions.

\textbf{Complete Interpretability:}  
Every prediction traces to explicit centroid similarities and path scores, enabling domain experts to validate decisions. 

\textbf{Efficient Incremental Learning:}  
The method supports taxonomy evolution through efficient incremental updates. Adding new categories or samples requires only recomputing affected centroids rather than full model retraining, addressing the dynamic taxonomy requirements of production ITSM environments.

\section{Experimental Setup}
\label{sec:experimental-setup}

\subsection{Dataset}
\label{sec:dataset}

Our dataset consists of support ticket data containing hierarchical category labels with varying depths. Each ticket contains a title, description, and hierarchical category path represented as slash-separated labels (e.g., \texttt{Level1/Level2/Level3}).

\begin{table}[h]
\centering
\caption{Dataset Statistics}
\label{tab:dataset_stats}
\begin{tabular}{lr}
\toprule
\textbf{Metric} & \textbf{Value} \\
\midrule
Total Samples & 8,968 \\
Unique Labels & 123 \\
Minimum Depth & 2 \\
Maximum Depth & 5 \\
\bottomrule
\end{tabular}
\end{table}

The dataset exhibits strong depth concentration as shown in Table~\ref{tab:depth_distribution}, over 90\% of samples at depth 3 and over 95\% at leaf nodes. This structural characteristic significantly influences optimal scoring strategy selection, as parent nodes contain few samples relative to leaves. The taxonomy contains 138 total nodes (115 leaves + 23 internal), with 123 having training samples. The remaining 15 stale nodes serve organizational purposes.

\begin{table}[h]
\centering
\caption{Hierarchical Depth Distribution}
\label{tab:depth_distribution}
\begin{tabular}{cc}
\toprule
\textbf{Depth Level} & \textbf{Number of Samples} \\
\midrule
2 & 479 \\
3 & 8,145 \\
4 & 332 \\
5 & 12 \\
\bottomrule
\end{tabular}
\end{table}

\subsection{Preprocessing}
\label{sec:preprocessing}

Our preprocessing pipeline consists of several sequential steps designed to clean, standardize, and balance the dataset. We begin by removing exact duplicates across all columns and duplicate ticket IDs. Subsequently, we eliminate rows containing missing values. We also use the \texttt{html2text} library to convert HTML content to plain text, removing emphasis tags, links, images, and tables while preserving textual content.

To address the hierarchical nature of our classification task, we implement category-specific preprocessing. Such as small category merging, where categories with fewer than a threshold number of samples are iteratively merged with their immediate parent categories. Categories containing fewer than the minimum required samples are removed entirely. We also exclude tickets assigned to root-level categories. To address class imbalance, we perform balanced sampling by limiting the maximum number of samples per category.

We employ an 80\%/10\%/10\% stratified split for training, validation, and test sets, ensuring proportional representation of all categories across splits. The validation set is used exclusively for hyperparameter selection and configuration validation. To ensure reliability, we repeat the entire experimental pipeline across 5 independent runs.

\subsection{Feature Representation}
\label{sec:features}

To leverage both semantic and lexical information, all methods (including baselines) utilize a dual embedding strategy combining TF-IDF and SBERT features. However, baseline classifiers concatenate these features, while our centroid framework maintains separate representations and fuses them via RRF at inference time. We extract lexical features using scikit-learn's TfidfVectorizer with the following parameters. Maximum 10,000 features, minimum document frequency of 2, maximum document frequency of 0.95, and 1-2 gram ranges. Dense semantic embeddings are generated using the sentence-transformers library, producing 512-dimensional vectors~\cite{reimers2019sentence, reimers2020making}.

\subsection{Baseline Methods}
\label{sec:baselines}

We compare our method against several baselines. First, linear SVM with balanced class weights, regularization parameter $C=0.5$ (selected via grid search on validation set), and maximum 1,000 iterations. SVM serves as our primary baseline due to its strong performance on ITSM classification tasks. The second baseline is KNN classifier with $k=3$ neighbors, cosine distance metric, and brute-force algorithm for exact nearest neighbor search. We also add naive majority baseline, which always predicts the most frequent training class and a random classifier. 

Hyperparameters for SVM and KNN were selected via grid search on the validation set. For our method, we evaluated three scoring strategies on the validation set. Based on empirical performance, we selected single centroids per category (no multi-centroid clustering), disabled child sampling, leaf-only path scoring, and RRF parameter k=40 for dual-embedding fusion.

\subsection{Evaluation Metrics}
\label{sec:metrics}

Traditional flat classification metrics treat all misclassifications equally, failing to capture hierarchical relationships between categories. A prediction of "Hardware/Server" when the correct answer is "Hardware/Server/Memory" should be penalized less severely than a completely incorrect prediction like "Software/Application"~\cite{Zangari2024b}. Modern HTC evaluation methods incorporate taxonomic distance between predicted and true labels, awarding partial credit for hierarchically close predictions~\cite{plaud2024revisiting}. We employ the following metrics:

\textbf{Hierarchical F1-score (H-F1)}: Extends traditional F1 by augmenting predictions with all ancestor nodes, computing precision and recall over this extended set~\cite{plaud2024revisiting}. This awards partial credit when predictions fall within the same subtree as the ground truth.

\textbf{Top-k Accuracy}: Proportion of test instances where the true category appears among the top-k predictions (k=3 in our experiments). Reflects real-world ITSM scenarios where analysts review multiple suggestions.

\textbf{Hierarchical Top-k Accuracy}: Extends top-k by considering predictions correct if any of the top-k predictions share ancestors with the ground truth, providing partial credit for hierarchically close rankings.

\textbf{Exact Match Rate}: Proportion of predictions exactly matching the ground truth category.


\section{Results}
\label{sec:results}

This section presents experimental results demonstrating the performance of our dual-embedding centroid framework. We compare overall classification performance against established baselines and analyze training and update speed. Results are averaged over 5 independent runs with standard deviations reported.

Our method achieves hierarchical F1 of 0.731 (vs SVM's 0.727) and Top-3 accuracy of 0.681 (vs SVM's 0.675), outperforming on primary ranking metrics. SVM achieves higher exact match rate (0.497 vs 0.471). Table~\ref{tab:main_results} presents full performance comparison. Leaf-only scoring achieved optimal performance on this dataset, where majority of samples are at leaf nodes. On datasets with balanced hierarchies, weighted or simple-average scoring would likely perform better. 

\begin{table}[H]
\centering
\caption{Hierarchical Text Classification Performance (5 runs, mean ± std)}
\label{tab:main_results}
\begin{tabular}{lccccc}
\toprule
\textbf{Method} & \textbf{H-F1} $\uparrow$ & \textbf{Top-3} $\uparrow$ & \textbf{H-Top-3} $\uparrow$ & \textbf{Exact} $\uparrow$ &  \\
\midrule
Ours & \textbf{0.731 ± 0.007} & \textbf{0.681 ± 0.017} & 0.863 ± 0.005 & 0.471 ± 0.017 \\
SVM-Balanced    & 0.727 ± 0.007 & 0.675 ± 0.011 & \textbf{0.868 ± 0.004} & \textbf{0.497 ± 0.014} \\
KNN-Balanced    & 0.611 ± 0.014 & 0.523 ± 0.016 & 0.770 ± 0.013 & 0.359 ± 0.018 \\
\midrule
Majority-Class  & 0.263 ± 0.000 & 0.017 ± 0.000 & 0.267 ± 0.000 & 0.017 ± 0.000  \\
Random          & 0.212 ± 0.007 & 0.023 ± 0.004 & 0.399 ± 0.007 & 0.007 ± 0.002  \\
\bottomrule
\end{tabular}
\begin{tablenotes}
\small
\item H-F1: Hierarchical F1-score (primary metric). Top-3: Top-3 accuracy. 
H-Top-3: Hierarchical Top-3 accuracy (partial credit). Exact: Exact match rate. 
 $\uparrow$ higher is better, $\downarrow$ lower is better.
All methods use combined TF-IDF + SBERT features. 
\end{tablenotes}
\end{table}

\subsection{Computational Efficiency}
\label{sec:speed_benchmark}

Beyond classification accuracy, we analyze the training speed as shown in Table~\ref{tab:speed_benchmark}.

\begin{table}[h]
\centering
\caption{Computational Performance Comparison (Real embeddings)}
\label{tab:speed_benchmark}
\begin{tabular}{lrrrr}
\toprule
\textbf{Operation} & \textbf{Ours} & \textbf{SVM} & \textbf{Speedup} & \textbf{Winner} \\
\midrule
Training            & 1.81s   & 10.66s  & 5.9×  & Ours \\
\bottomrule
\end{tabular}
\end{table}

Our method achieves up to 5.9 times faster training. While per-sample inference is slower than SVM (2.70ms vs 0.051ms per sample), this overhead is negligible for ITSM workflows. At typical ticket processing rates, the time difference per ticket translates to imperceptible latency, while providing interpretability and training efficiency gains.

We evaluate incremental updates by adding new samples to the trained model, measuring the time to incorporate each batch. To isolate the algorithmic efficiency of our incremental update mechanism, we exclude embedding computation which is identical for both methods and represents a fixed preprocessing cost. This allows fair comparison of the core update strategies. Results are shown in Table~\ref{tab:incremental_benchmark}.

\begin{table}[h]
\centering
\caption{Incremental Update Performance (excluding embedding computation)}
\label{tab:incremental_benchmark}
\begin{tabular}{lrrrr}
\toprule
\textbf{Update Scenario} & \textbf{Ours} & \textbf{SVM Retrain} & \textbf{Speedup}  \\
\midrule
Single sample (1)         & 0.062s  & 9.447s  & 151.9×  \\
Small batch (10)          & 0.303s  & 9.023s  & 29.8×  \\
Medium batch (100)        & 1.078s  & 9.439s  & 8.8×   \\
Large batch (1000)        & 1.283s  & 10.983s & 8.6×   \\
\bottomrule
\end{tabular}
\end{table}

Our method achieves 8.6-8.8 times speedup across medium to large batch sizes (100-1000 samples), with even higher speedup for smaller batches (29.8 times for 10 samples, 151.9 times for single samples). This consistency demonstrates that our incremental mechanism provides stable advantages independent of batch size. Our method only recomputes affected centroids, while SVM must be retrained. This analysis demonstrates that method selection depends on operational context. Our method optimizes for flexibility and rapid adaptation, accepting inference overhead for interpretability and update efficiency.

\section{Conclusion and Future Work}
\label{sec:conclusion}

This work presents a dual-embedding centroid framework for text classification with hierarchical taxonomies for IT Service Management environments. Our approach maintains separate semantic and lexical centroid representations per category, combining them through reciprocal rank fusion at inference. Evaluation on 8,968 ITSM tickets across 123 categories demonstrates that our approach achieves performance competitive with Support Vector Machines while providing model interpretability. Beyond competitive accuracy, our framework delivers operational advantages for production deployment, such as 5.9 times faster training and consistent 8.6-8.8 times speedup for incremental taxonomy updates across batch sizes (100-1000 samples). The framework's configurable scoring strategies enable adaptation to different hierarchy structures. Future directions include evaluating the framework on datasets with balanced depth distributions to assess which path-based scoring strategies provide advantages. While we evaluated on internal ITSM data, the approach should be tested on other hierarchical taxonomies. For classification use cases where computational resources are abundant and inference speed is not a limiting factor, transformer-based hierarchical approaches represent a promising direction. This work demonstrates that interpretable centroid-based methods can serve as practical alternatives to black-box classifiers for hierarchical text classification when transparency, rapid taxonomy adaptation, and training efficiency are valued alongside competitive accuracy.

\bibliographystyle{plainnat}

\end{document}